\documentclass[letterpaper, 10 pt, conference]{ieeeconf}  

\IEEEoverridecommandlockouts                              

\usepackage{fancyhdr}
\usepackage{graphicx}
\usepackage{subcaption} 
\usepackage{wrapfig}
\usepackage{cite}
\usepackage{picinpar}
\usepackage{amsmath}
\usepackage{amssymb}
\usepackage{amsfonts}
\usepackage{url}
\usepackage[latin1]{inputenc}
\usepackage{colortbl}
\usepackage{soul}
\usepackage{multirow}
\usepackage{multicol}
\usepackage{booktabs}
\usepackage{pifont}
\usepackage{color}
\usepackage{alltt}
\usepackage{dsfont}
\usepackage[hidelinks]{hyperref}
\usepackage{enumerate}
\usepackage{siunitx}
\usepackage{epstopdf}
\usepackage{pbox}
\usepackage[ruled, linesnumbered]{algorithm2e}
\usepackage{xr}
\usepackage{float}
\usepackage{tabularx}
\usepackage{array}
\usepackage{diagbox}

\title{\LARGE \bf
WIT-UAS: A Wildland-fire Infrared Thermal Dataset \\ to Detect Crew Assets From Aerial Views
}

\author{Andrew Jong$^{1}$, Mukai Yu$^{1}$, Devansh Dhrafani$^{2}$, Siva Kailas$^{1}$, Brady Moon$^{1}$, \\ Katia Sycara$^{1}$, Sebastian Scherer$^{1}$
    \thanks{$^{1}$Andrew Jong, Mukai Yu, Siva Kailas, Brady Moon, Katia Sycara, and Sebastian Scherer are with Carnegie Mellon University, School of Computer Science, the Robotics Institute, 5000 Forbes Ave, Pittsburgh, PA, USA.  (\href{mailto:ajong@andrew.cmu.edu}{ajong}, 
    \href{mailto:mukaiy@andrew.cmu.edu}{mukaiy},
    \href{mailto:skailas@andrew.cmu.edu}{skailas}, 
    \href{mailto:bradym@andrew.cmu.edu}{bradym}, 
    \href{mailto:sycara@andrew.cmu.edu}{sycara}, 
    \href{mailto:basti@andrew.cmu.edu}{basti}\}@andrew.cmu.edu)}
    \thanks{$^{2}$Devansh Dhrafani is with Carnegie Mellon University, College of Engineering, Department of Mechanical Engineering, 5000 Forbes Ave, Pittsburgh, PA, USA. (\href{mailto:ddhrafan@andrew.cmu.edu}{ddhrafan@andrew.cmu.edu})}
}%

\begin{document}

\maketitle
\thispagestyle{empty}
\pagestyle{empty}

\begin{abstract}

We present the Wildland-fire Infrared Thermal (WIT-UAS) dataset for long-wave infrared sensing of crew and vehicle assets amidst prescribed wildland fire environments.
While such a dataset is crucial for safety monitoring in wildland fire applications, to the authors' awareness, no such dataset focusing on assets near fire is publicly available.
Presumably, this is due to the barrier to entry of collaborating with fire management personnel.
We present two related data subsets: WIT-UAS-ROS consists of full ROS bag files containing sensor and robot data of UAS flight over the fire, and WIT-UAS-Image contains hand-labeled long-wave infrared (LWIR) images extracted from WIT-UAS-ROS.
Our dataset is the first to focus on asset detection in a wildland fire environment. 
We show that thermal detection models trained without fire data frequently detect false positives by classifying fire as people. 
By adding our dataset to training,  we show that the false positive rate is reduced significantly.
Yet asset detection in wildland fire environments is still significantly more challenging than detection in urban environments, due to dense obscuring trees, greater heat variation, and overbearing thermal signal of the fire.
We publicize this dataset to encourage the community to study more advanced models to tackle this challenging environment.
The dataset, code and pretrained models are available at \url{https://github.com/castacks/WIT-UAS-Dataset}.

\end{abstract}

\section{INTRODUCTION}
Wildfires today grow increasingly more drastic due to
climate extremification. As wildfires grow in
size and complexity, crews on the front lines need access to
greater situational awareness of fire position relative to themselves for
safety. Unmanned aerial systems (UAS) offer an ideal platform to provide such situational awareness, providing an eye in the sky while
being relatively cost-effective and portable. Greater levels of
autonomy for unmanned aerial systems would make this tool
easier to adopt at scale, requiring less training for human pilots
and providing greater capability to monitor the environment.
Wildfires provide a rich environment for research in robotics
and autonomy due to the complexity of fast dynamics, partial
observability, and risk in this dangerous environment.

While several works focus on estimating fire position \cite{Giuseppi2021UAVPF, castagno2021multi, shobeiry2021uav, Aggarwal2020MultiUAVPP}, none seek to perceive fire crew and vehicle assets.
The assumption that asset detection systems trained on data from nominal environments could translate seamlessly to wildfire application is naive since wildland fire environments present unique complexities for perception.
When observing large fires, RGB cameras are susceptible to smoke and are often unable to directly identify fire locations.
Even though thermal cameras penetrate smoke, detectors must overcome the overbearing heat signature from the fire itself.
Indeed, our experiments that test a neural detector trained on non-fire data (HIT-UAV \cite{hituav}) show all fire falsely detected as objects of interest, as illustrated in Figure \ref{fig:ssd_false_positive_fire_comparision}.
Another major discrepancy between existing, often urban-based, datasets and wilderness environments is the terrain.
Dense vegetation often obscures visibility in a way not found amongst human infrastructure.
Overall, there is a significant gap between existing datasets and the wildland fire environment, which causes systems trained on current datasets to have trouble generalizing for wildland fire applications.

\begin{figure}[t]
    \centering
    \includegraphics[width=0.48\textwidth]{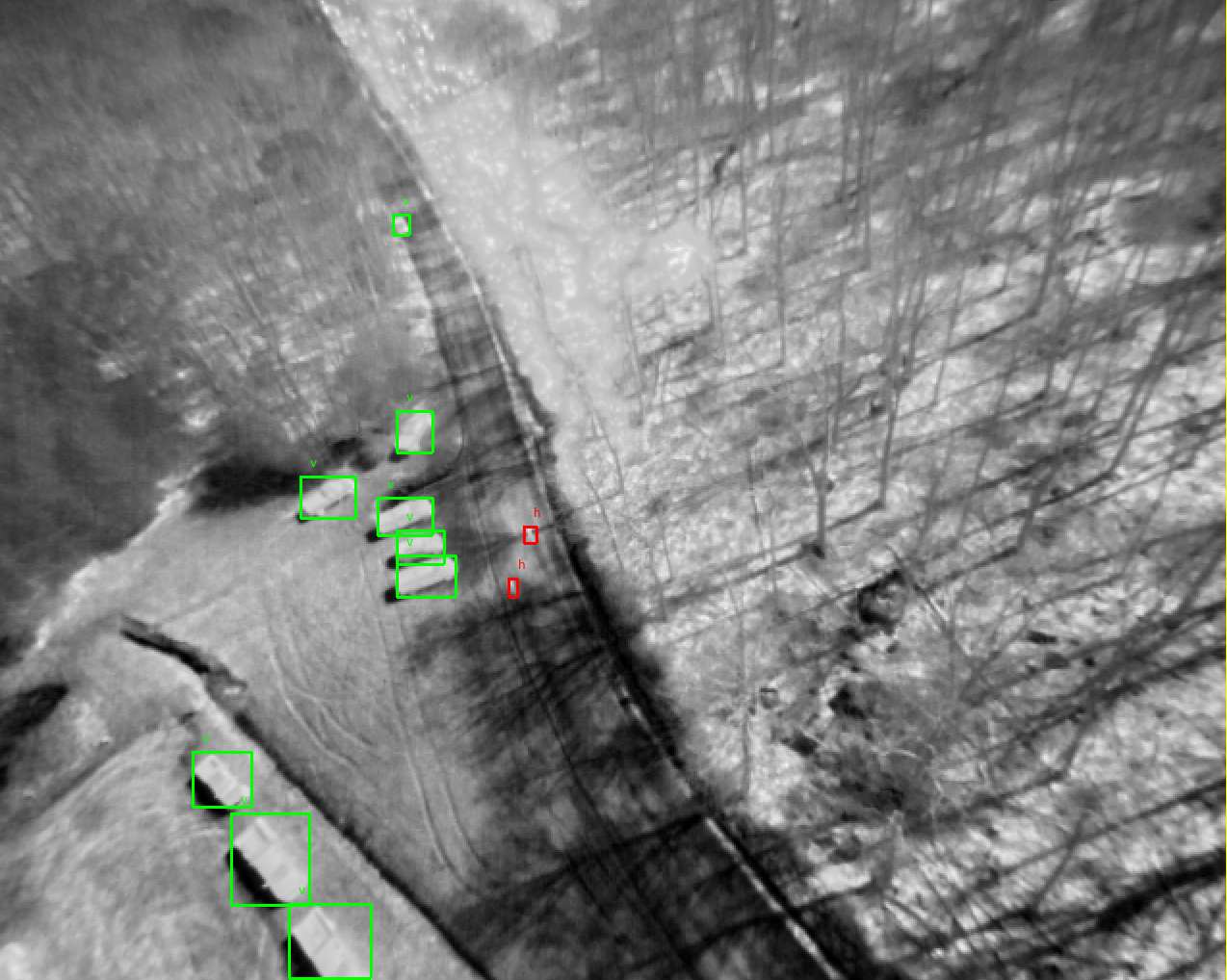}
    \caption{\small{Labeled long-wave infrared (LWIR) image from our WIT-UAS-Image dataset features vehicles (green, marked "v'') and people (red, marked "h'' for "human'') next to a wildland prescribed fire. The fire is the glowing heat in the middle of the picture, right of the road. The camera is pitched 30\textdegree\ down from the horizontal.}}
    \label{fig:labeled_sample_thermal_image}
\end{figure}

More data is needed to bridge the gap for robust asset detection for wildland fires.
However, collecting useful data is challenging.
First, wildland fire locations become strictly controlled zones to ensure public safety, which limits access.
Large wildfires are especially restricted to the public due to their dangerous intensity and unpredictable nature.
Aside from wildfires, seasonal prescribed burns provide a reasonable proxy source for fire data.
Controlled prescribed burns are used to reduce the build-up of fuels from dead vegetation, thus reducing the intensity and likelihood of sparking extreme uncontrolled wildfires.
Though prescribed burns yield lower-intensity flames, it is a close replication of wildfires.
Yet prescribed burns present their own set of challenges, as they may only take place in locations far from urban centers and only during the spring and fall under specific wind and humidity conditions.
All this explains the scarcity of such datasets.

Hence we present the Wildland-fire Infrared Thermal Unmanned Aerial System dataset (WIT-UAS), an aerial remote-sensor dataset that we collected over three seasons of controlled prescribed burns in western Pennsylvania.
All burns yield labeled thermal sensor data of people and vehicle assets.
To make the data useful for developing sensor models for downstream planning algorithms, we include sensor pitch and altitude information by providing full Robot Operating System (ROS) bags.
We expect this data to be useful in furthering robotics autonomy in wildland fires.

In this manuscript, we first summarize the robotics literature around wildland fire applications, as well as existing thermal and UAV datasets.
Next, we describe the data collection setup and labeling process for WIT-UAS.
We analyze the data to illustrate its quantitative distribution and qualitative properties.
Next, we document our examples of training standard neural detection models on this data and analyze the results. 
Our results show that this unique environment presents a significant challenge for detection performance, and conveys the necessity to develop more sophisticated models.
We conclude with recommendations for future work towards the ultimate goal of an autonomous wildland fire monitoring system.


\section{RELATED WORK}
\textbf{Wildland Fire UAS: }
The hazardous and visually degraded environment makes wildfire monitoring challenging, and manned aerial monitoring of assets (crew, vehicles, etc) in wildfire situations is both expensive and risky, especially if there are uncontrolled fires \cite{bailonsurvey}. Autonomous UAV systems such as \cite{bailon2020design} and \cite{Esmaeil} have been proposed, but generally abstract the perception aspect of autonomy. Since the perception problem for tracking assets in wildland fire monitoring is difficult, it warrants a solution on its own in order to provide robust detection and tracking. The development of such perception systems would enable UAV systems such as those proposed by \cite{bailon2020design} and \cite{Esmaeil} to be utilized in real-world scenarios. To address providing data to enable machine learning approaches, this dataset was curated in hopes of enabling novel perception systems for UAV monitoring of assets in wildland fires.


\textbf{Object Detection: }
Object detection has always been a prevalent task within the field of computer vision.
As a result, object detection datasets have been integral in enabling methodologies such as deep learning and supervised learning in general to have more robust performance.
For example, MSCOCO \cite{mscoco}, ImageNet \cite{imagenet}, and Pascal VOC \cite{everingham2009pascal} are three general object detection datasets that are widely used as benchmark datasets for general object detection.
These datasets have given rise to novel model architectures such as SSD \cite{Liu_2016} and YOLO \cite{yolo, yolo9000, yolov3} for which pre-trained models now exist.
Both of the detectors utilize a single-stage network architecture design, which is beneficial for computationally limited UAV platforms.

In the context of object detection for UAVs, there have been attempts to tackle small object detection, e.g., UAV-YOLO \cite{UAV-YOLO}, which employs multi-scale receptive field in its network design; and SAHI \cite{SAHI}, which is an inference framework on top of existing pre-trained object detector that extracts bounding boxes from crops of the entire image then merges the boxes together afterward.
However, most of the methods cannot enable small object detection without introducing additional overhead on computation, which is especially limited on UAV platforms.
The WIT-UAS targets more on heterogeneous sensing modality and scene setting, thus the impact of object scale on detectors is not extensively discussed in this paper.

However, in order to leverage novel deep learning architectures or supervised learning methodologies in other domains, robust datasets often need to be procured for the domain of interest.
Towards the use of UAVs in domains such as forest fire prevention, recent object detection datasets have utilized UAVs and various sensor modalities such as RGB or thermal cameras.

\textbf{UAV RGB or Visible Datasets: }
There are many UAV-based datasets for RGB or visible spectrum images. Some examples include DTB \cite{Li_Yeung_2017}, UAV123 \cite{mueller2016benchmark}, and UAVDT \cite{du2018unmanned}. UAV123 uses a mixture of real and synthetic data, whereas DTB and UAVDT are only composed of real images. Moreover, UAVDT contains annotations for detection and multiple object tracking, whereas DTB and UAV123 only contain labels for detection alone. UAV123 has the largest size of data, with 113K images, while UAVDT has 80K images and UTB has 13K images.

\textbf{UAV Thermal Datasets: }
While RGB and visual data are the primary sensor modalities for general object detection, such data is inherently at a disadvantage for low-light scenarios. Scenarios such as wildfire monitoring, search-and-rescue, or disaster relief can occur either during the day or at night. Thus, the use of thermal infrared sensors has become preferable for datasets in these domains since they can capture
images in both day and night scenarios. Bondi et al. procured a dataset aimed at using UAVs to deter animal and wildlife poaching by collecting a mixture of real and synthetic thermal images of humans and animals \cite{birdai}. Suo et al. proposed a dataset of real thermal images collected from high-altitude UAVs for urban objects (such as people, vehicles, bicycles, etc) \cite{hituav} called HIT-UAV. Both of these, while relevant to the domain of wildfire monitoring, lack the presence of fire, which we empirically show has a tremendous effect on the inference ability.

\textbf{UAV Fire Datasets: }
One notable dataset that includes aerial images of fires is the FLAME dataset \cite{shamsoshoara2021aerial}. This dataset offers both visual and thermal images and videos with the intention of aiding deep-learning methodologies for fire detection. However, such a dataset would not be as useful for the task of object detection since there are no objects present. This motivates the WIT-UAS dataset.


\section{WIT-UAS Dataset}

\subsection{Data Collection Setup}
The WIT-UAS-ROS ROS bag files were collected from three prescribed fire seasons (fall 2021, spring 2022 and fall 2022).
Prescribed burns were set and managed by the Pennsylvania Game Commission at various State Game Lands (SGL) in Western Pennsylvania, namely (1) SGL 174 near Rossiter, PA, (2) SGL 111 near Confluence, PA, and (3) SGL 42 Reade Township, PA. The authors worked with the Game Commission to secure a permit to operate UAVs over prescribed fires.

The UAV platforms used are the DJI M100 and DJI M600. The DJI M100 was used for SGL 174, whereas the DJI M600 was used for SGL 111 and SGL 42. The UAVs were mounted with an onboard NVIDIA Jetson Xavier NX computer that was connected to the sensors.

\begin{figure}[ht]
    \centering
    \includegraphics[height=.23\textwidth]{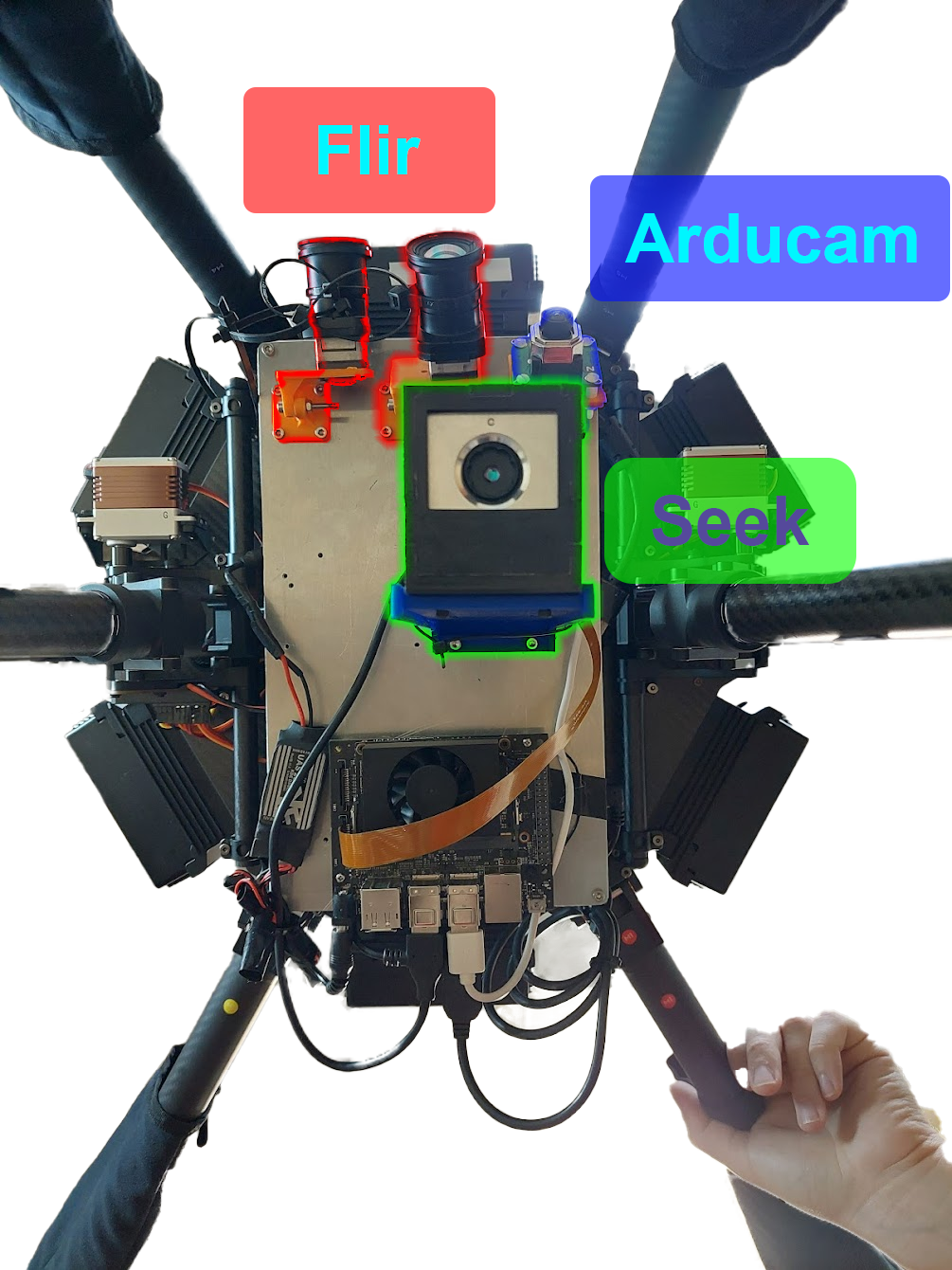}
    \includegraphics[height=.23\textwidth]{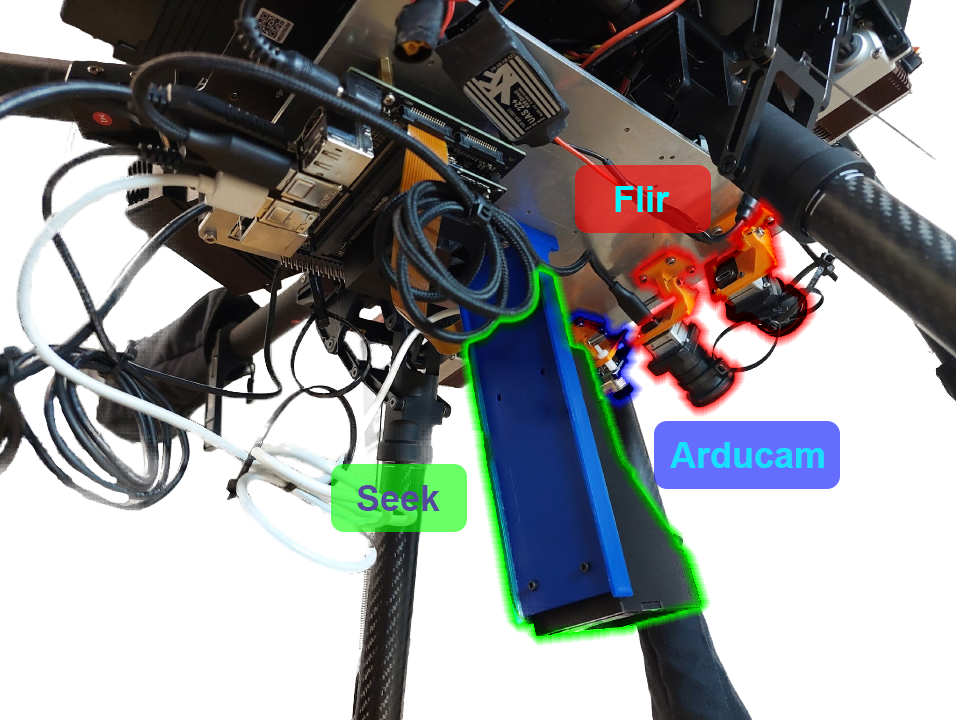}
    \caption{\small{DJI M600 hardware setup flown over SGL 111 near Confluence, PA. Two FLIR cameras are forward-mounted pitched down 30\textdegree\; from the horizontal. The Seek camera is mounted under the blue plastic facing straight down.}}
    \label{fig:drone_setup}
\end{figure}

As the data collection platform developed, different sensors were added or removed. 
All ROS bags include data reported by the DJI ROS SDK, including sources for odometry (GPS, IMU), drone state (pose, velocity), and telemetry (battery, GPS health, etc).
The ROS bags then include recorded camera sensor data.
SGL 174 near Rossiter, PA featured one FLIR Boson 640 95\textdegree HFOV 4.9mm long-wave infrared (LWIR) camera mounted 90\textdegree down from the horizontal.
SGL 111 near Confluence, PA includes three LWIR cameras: one absolute-temperature Seek S304SP 56\textdegree HFOV mounted 90\textdegree\ down from the horizontal with $320\times240$ resolution, one relative-temperature FLIR Boson 50\textdegree HFOV pitched 30\textdegree\ down with $640\times512$ resolution, and one relative-temperature FLIR Boson 95\textdegree HFOV pitched 30\textdegree\ down. This is shown in Figure \ref{fig:drone_setup}.
SGL 42 near Reade Township, PA contained the same Seek camera configuration as from SGL 111 and removed the two FLIR cameras.
However, SGL 42 also added one $1920\times1080$ resolution RGB camera (Arducam IMX477) at a downward 90\textdegree\ angle.
Table \ref{tab:uas-data} neatly summarizes the above information.
UAVs were manually piloted within visual line of sight to fly over fire and assets.
All pilots were Part 107 certified and trained for UAS flight according to lab procedure.

Sensor data was recorded through the Robot Operating System (ROS) running onboard.
Sensors were synchronized through ROS time (not hardware), making WIT-UAS unsuitable for precise mapping, but still useful to study approximate vantages from high altitudes.
Approximate position information is useful to develop sensor models, say for an autonomous planner that sets waypoints to track assets with a desired confidence.

\begin{table}[ht]
\caption{UAS data collection information}
\centering
    \begin{tabular}{|c|c|c|c|}
        \hline
        \textbf{Burn Site} & Rossiter & Confluence & Reade \\ \hline
        \textbf{Season} & Fall 2021 & Spring 2022 & Fall 2022 \\ \hline
        \textbf{UAV} & DJI M100 & DJI M600 & DJI M600 \\ \hline
        \textbf{Sensor} & FLIR & FLIR/Seek & Seek/Arducam \\ \hline
        \textbf{Sensor Type} & Thermal & Thermal/Thermal & Thermal/RGB \\ \hline
        \textbf{Sensor Pitch} & 90\textdegree & 30\textdegree/90\textdegree & 90\textdegree/90\textdegree \\ \hline
        \textbf{\# Labeled Images} & 1033 & 2935/2542 & 570/- \\ \hline
    \end{tabular}
\label{tab:uas-data}
\end{table}

\begin{figure}[!ht]
    \centering
    \begin{subfigure}[b]{0.22\textwidth}
        \includegraphics[width=\textwidth]{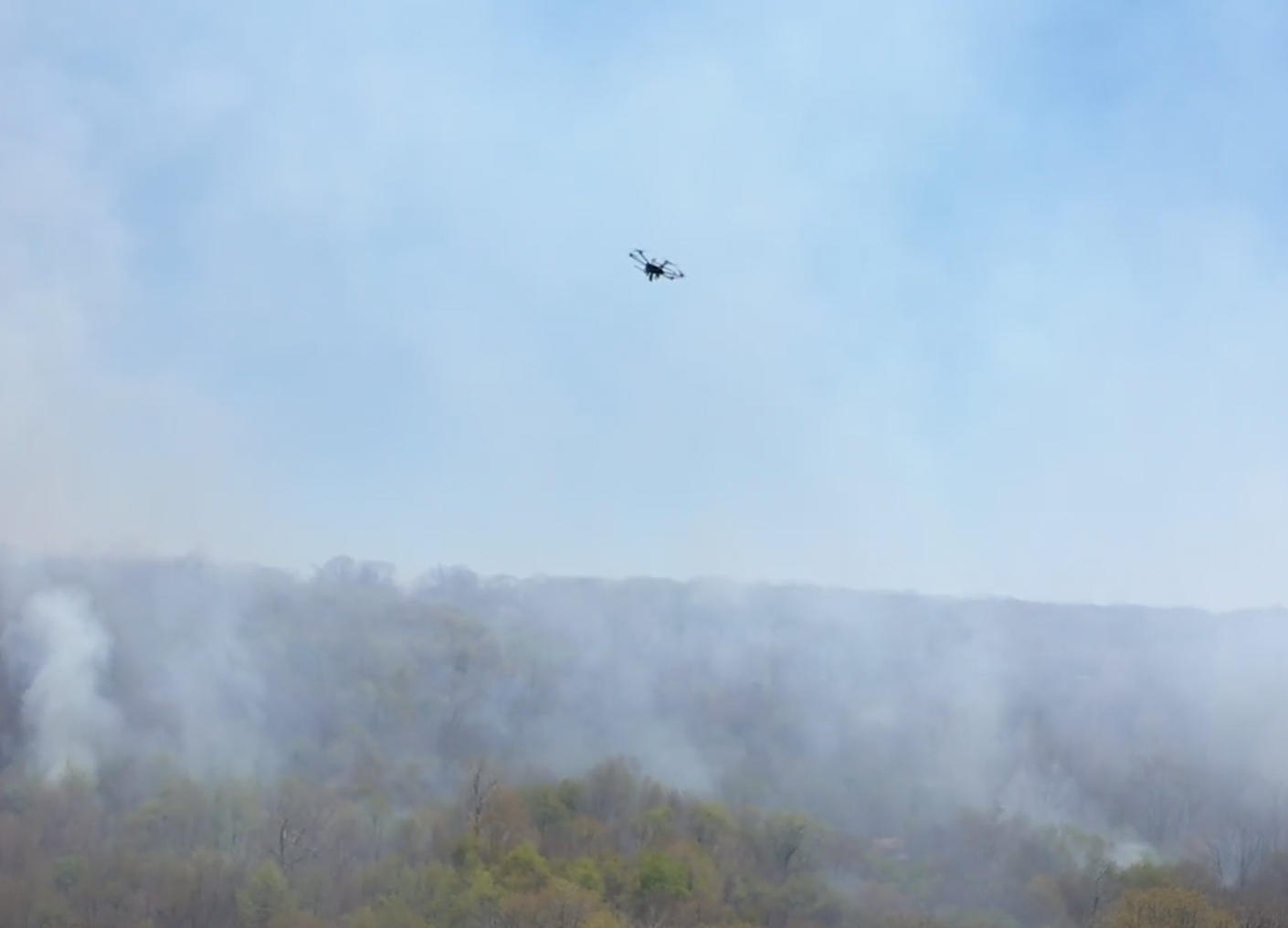}
        \caption{SGL 111 Confluence}
    \end{subfigure}
    \begin{subfigure}[b]{0.24\textwidth}
        \includegraphics[width=\textwidth]{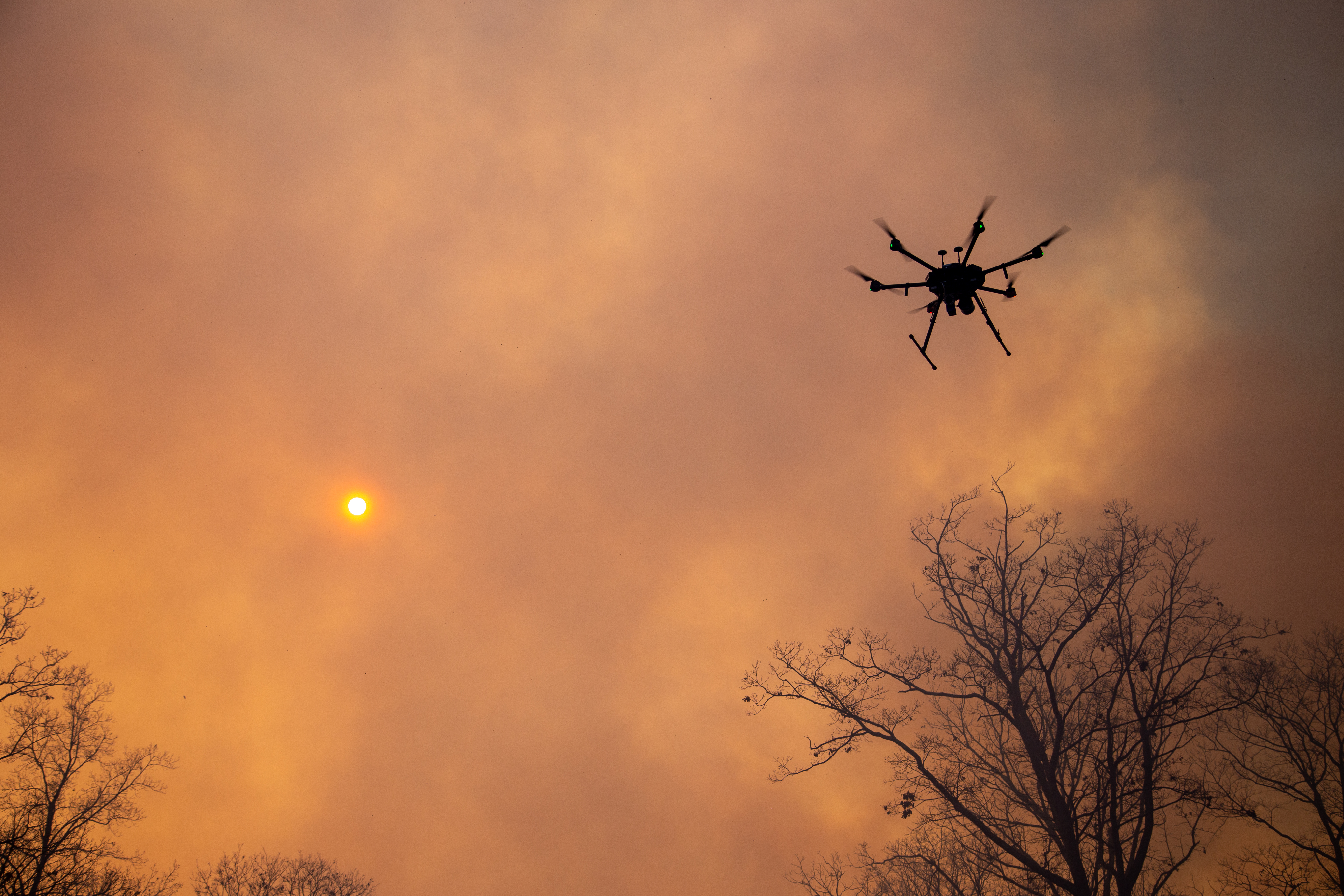}
        \caption{SGL 42 Reade Township}
    \end{subfigure}
    \caption{\small{Dataset collection environment. Both images were taken on days with clear blue skies around mid-day. The image from Reade Township looks tinted orange because smoke covers the sun.}}
    \label{fig:dataset_collection_env}
\end{figure}

\subsection{Data Processing and Labeling}

Since thermal is the common sensor type throughout all three collection sites and yielded the highest quality data for object detection due to its smoke penetration capability, we produce the labeled WIT-UAS-Image bounding box dataset solely from the thermal images.
Two thermal sensors are featured in the WIT-UAS image dataset.
First is the FLIR Boson 640 95\textdegree  HFOV 4.9 mm long-wave infrared camera that has a $640\times512$ image resolution.
Second is the Seek S304SP 56\textdegree HFOV 4.0 mm long-wave infrared camera that has a $320\times240$ image resolution.
From the recorded ROS bags,
we extract images to label at 1 frame per second, resulting in 6951 total images.
We labeled the two most important assets of interest to wildland fighters: humans and vehicles.
The labelers were either knowledge experts present at the burn site or instructed by such experts on how to properly identify assets.
Labelers were instructed to label assets if they could identify them using whatever context cues available; for example, by using contextual knowledge that fire setters would lead the firefront, or to use motion cues from viewing images in sequence.
Labeling was performed using custom lab-developed auto-labeling software, which is open source and may be found online\footnote{\href{https://bitbucket.org/castacks/automatic-frame-labelling}{https://bitbucket.org/castacks/automatic-frame-labelling}}.
The auto-labeling works by tracking the anchor of object bounding boxes in a sequence of temporally adjacent images given manually labeled objects in the first frame.
The software produced \texttt{.label} files with bounding boxes described in Pascal VOC format, listed per line as minX, minY, maxX, maxY, and class.
However, the tracking performance of the auto-labeling was sub-optimal since the spectrum of noise introduced by raw thermal images is disparate compared to RGB images.

\begin{figure*}[ht!]
    \centering
    \begin{subfigure}[b]{0.24\textwidth}
        \includegraphics[width=\textwidth]{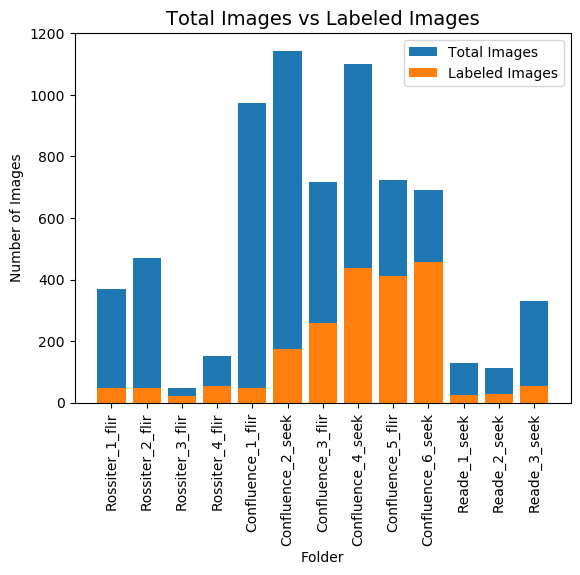}
    \end{subfigure}
    \begin{subfigure}[b]{0.24\textwidth}
        \includegraphics[width=\textwidth]{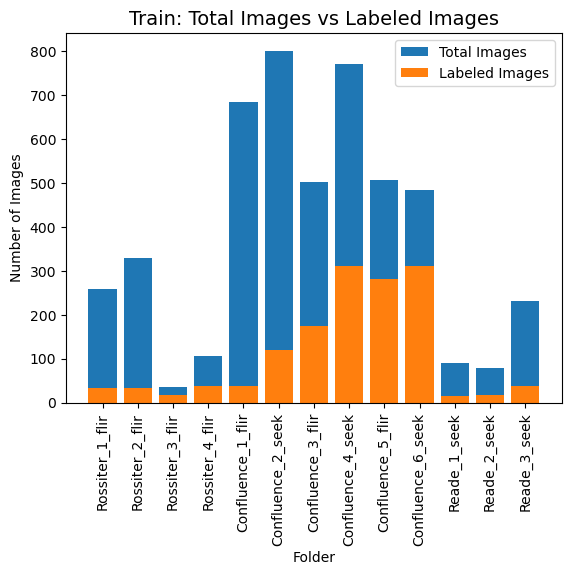}
    \end{subfigure}
    \begin{subfigure}[b]{0.24\textwidth}
        \includegraphics[width=\textwidth]{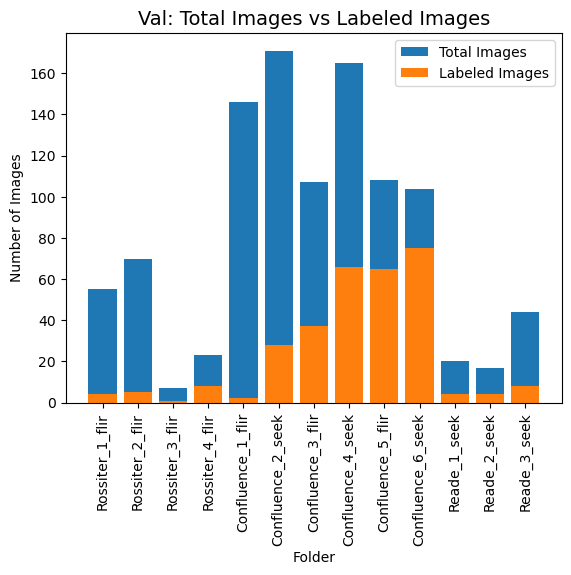}
    \end{subfigure}
    \begin{subfigure}[b]{0.24\textwidth}
        \includegraphics[width=\textwidth]{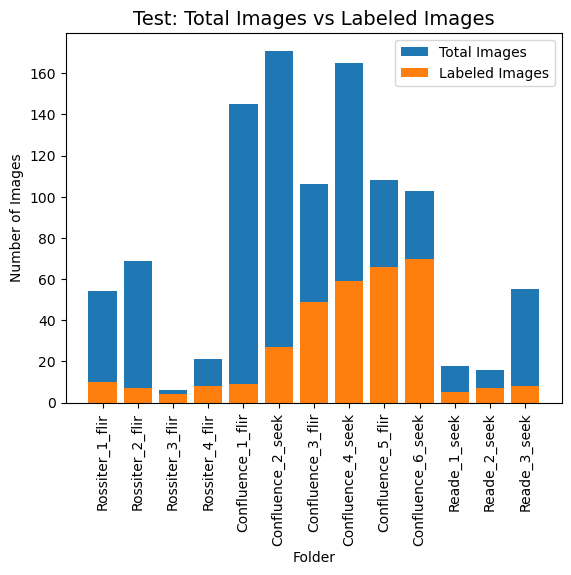}
    \end{subfigure}
    \\
    \begin{subfigure}[b]{0.24\textwidth}
        \includegraphics[width=\textwidth]{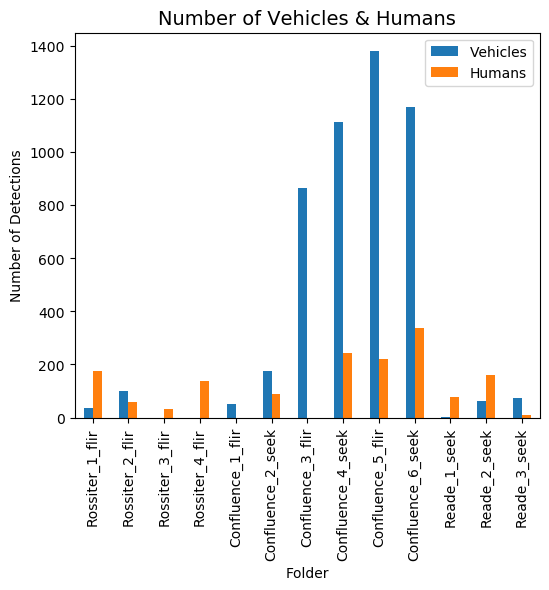}
    \end{subfigure}
    \begin{subfigure}[b]{0.24\textwidth}
        \includegraphics[width=\textwidth]{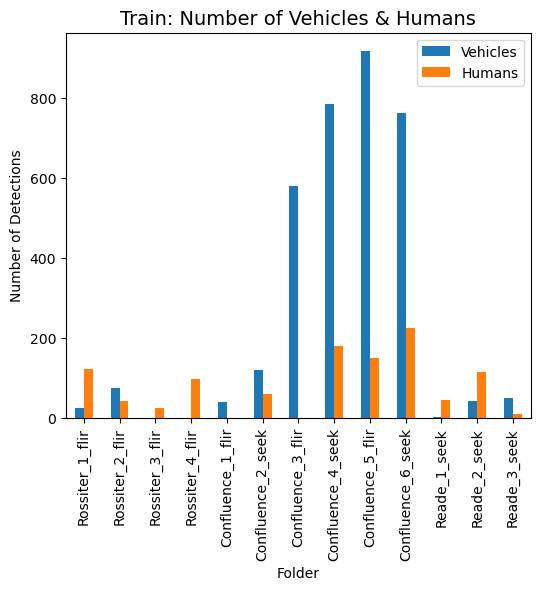}
    \end{subfigure}
    \begin{subfigure}[b]{0.24\textwidth}
        \includegraphics[width=\textwidth]{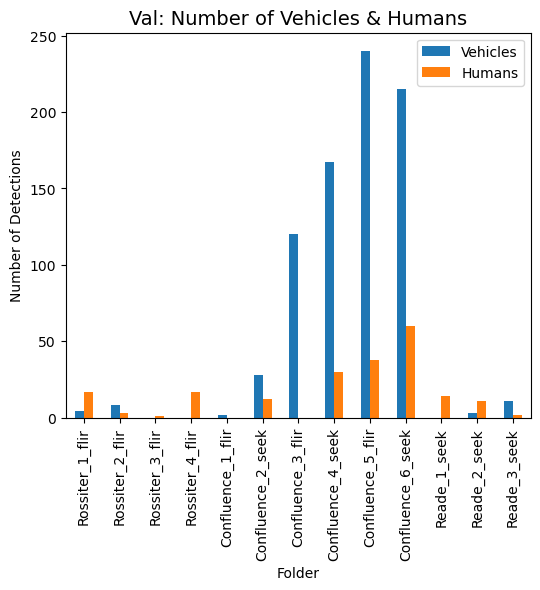}
    \end{subfigure}
    \begin{subfigure}[b]{0.24\textwidth}
        \includegraphics[width=\textwidth]{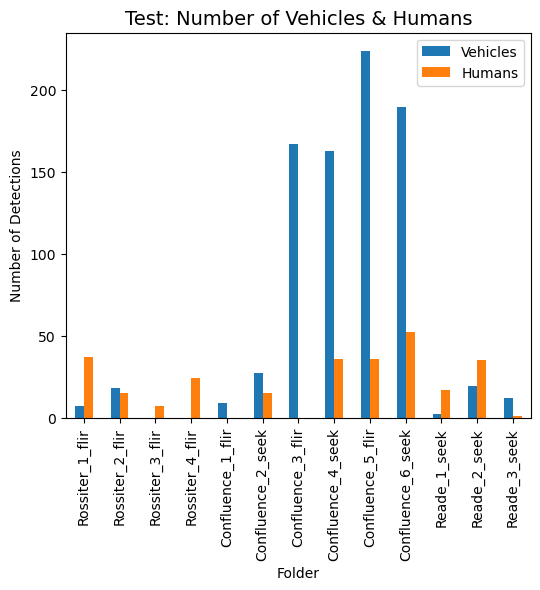}
    \end{subfigure}
    \caption{\small{Relative distributions of labeled images and object classes between the cumulative dataset (first column), train subset (second column), validation subset (third column), and test subset (fourth column). Top row: number labeled vs unlabeled images. Bottom row: number of vehicles vs number of humans.}}
    \label{fig:data_distribution}
\end{figure*}

After labeling all the images, we divide the data into train, validation, and test subsets, with approximately a 70\%, 15\%, 15\% split, resulting in 4877, 1037, and 1037 images respectively.
We desired each subset to have similar distributions of the number of objects and environment.
Therefore, the split was created by grouping the dataset into groups of ten, and randomly selecting either one or two images to belong to the validation or test set, and the remaining to be part of the train set.
This random selection resulted in similar distributions of labeled objects, and ensured similar pixel distribution in images, as shown in Figure \ref{fig:data_distribution}
We later evaluate generalization by evaluating out-of-distribution data from HIT-UAV.

\begin{figure}[t]
    \centering
    \begin{subfigure}[b]{0.23\textwidth}
        \includegraphics[width=\textwidth]{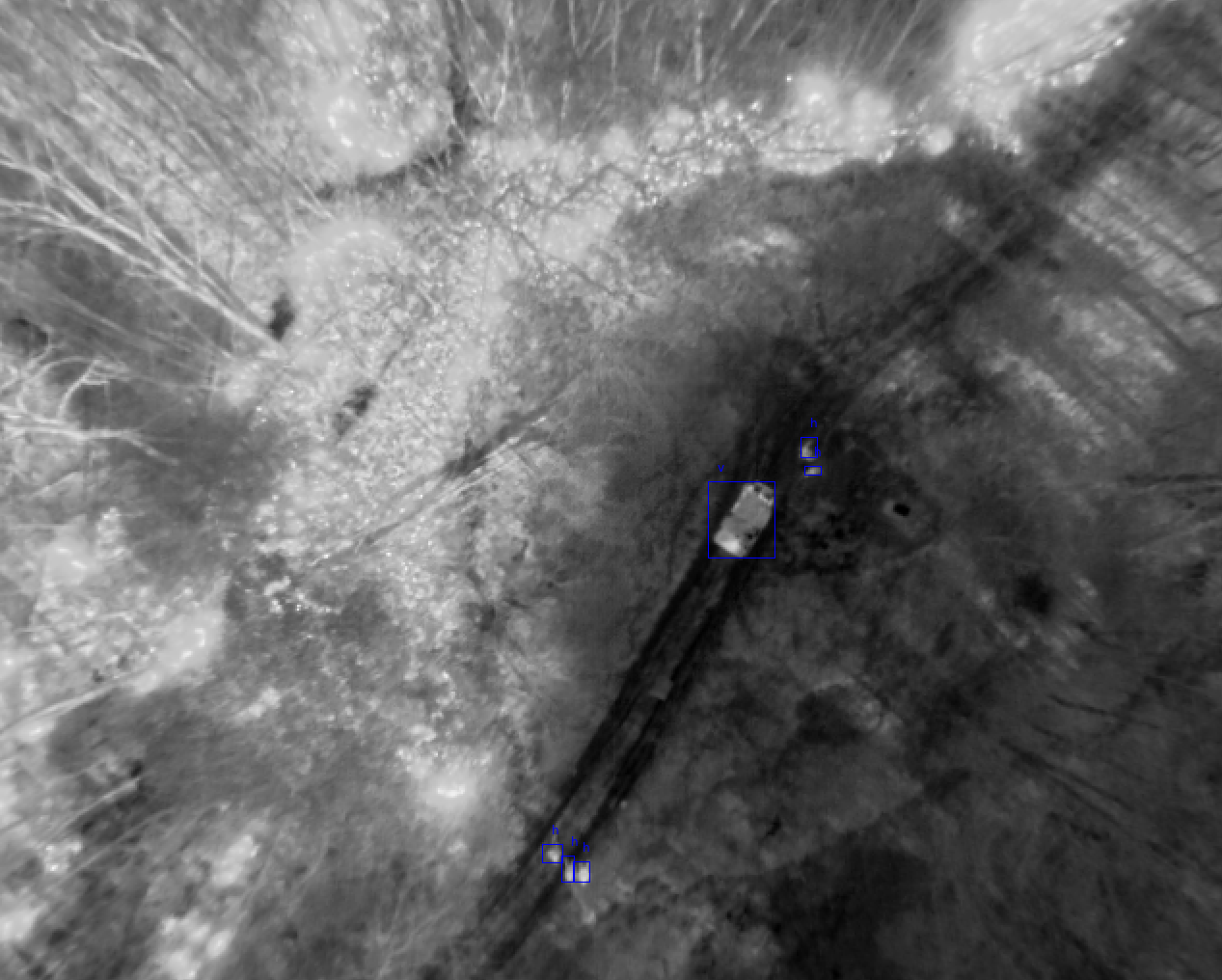}
        \caption{SGL 174 Rossiter, downward FLIR Boson $640\times512$}
    \end{subfigure}
    \begin{subfigure}[b]{0.23\textwidth}
        \includegraphics[width=\textwidth]{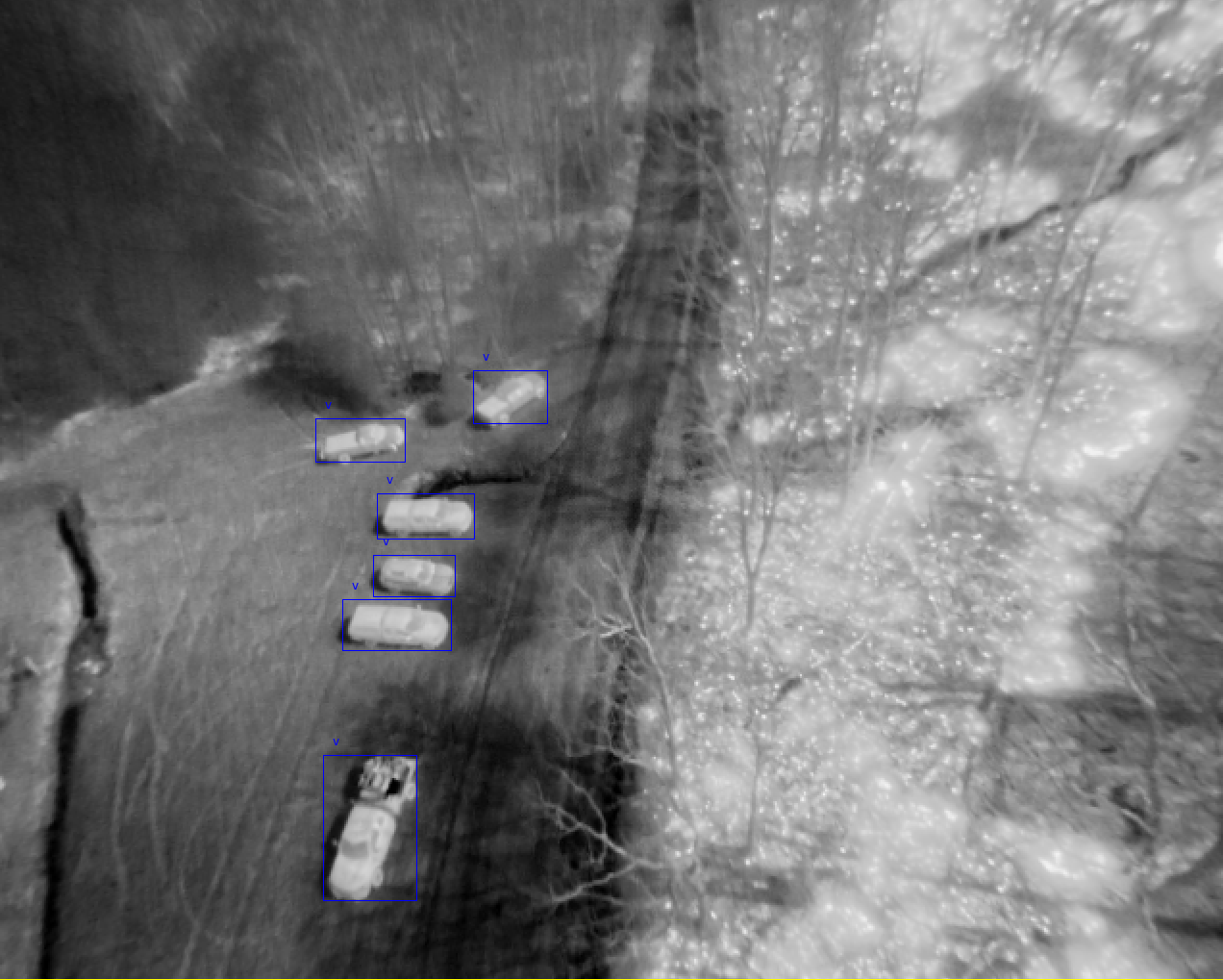}
        \caption{SGL 111 Confluence, 30\textdegree\; pitched FLIR Boson $640\times512$}
    \end{subfigure}
    \begin{subfigure}[b]{0.23\textwidth}
        \includegraphics[width=\textwidth]{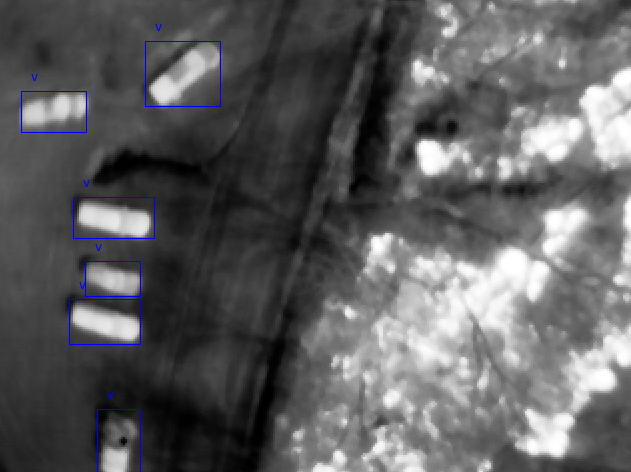}
        \caption{SGL 111 Confluence, downward Seek Camera $320\times240$}
    \end{subfigure}
    \begin{subfigure}[b]{0.23\textwidth}
        \includegraphics[width=\textwidth]{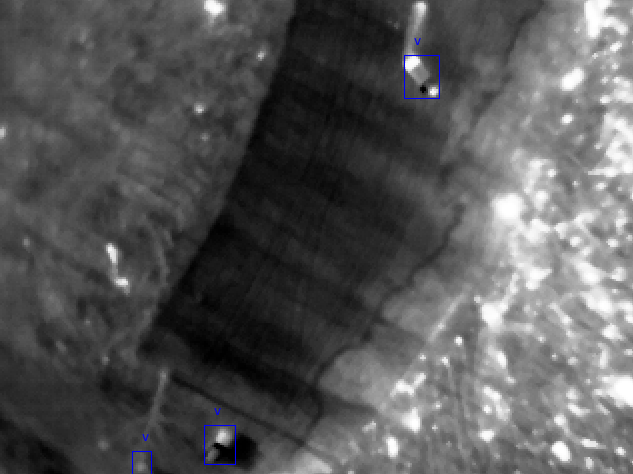}
        \caption{SGL 42 Reade,  downward Seek Camera $320\times240$}
    \end{subfigure}
    
    \caption{\small{Samples labeled thermal LWIR images from our presented WIT-UAS-Image dataset. Samples showcase different pitch angles and sensor types.}}
    \label{fig:sample_thermal_images}
\end{figure}

\begin{figure}[t]
    \centering
    \begin{subfigure}[b]{0.23\textwidth}
        \includegraphics[width=\textwidth]{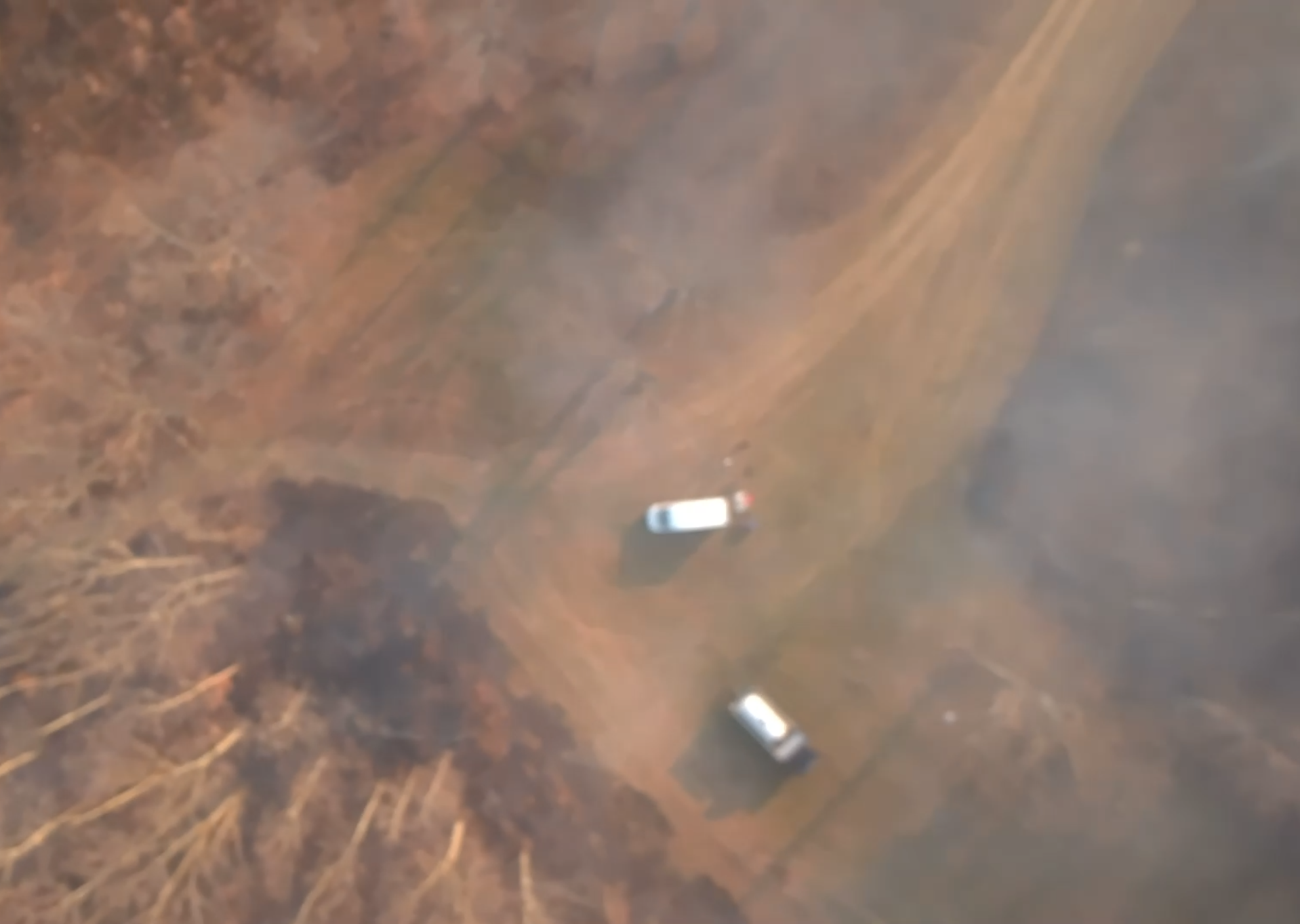}
        \caption{Visible vehicle and people assets in light smoke.}
    \end{subfigure}
    \begin{subfigure}[b]{0.23\textwidth}
        \includegraphics[width=\textwidth]{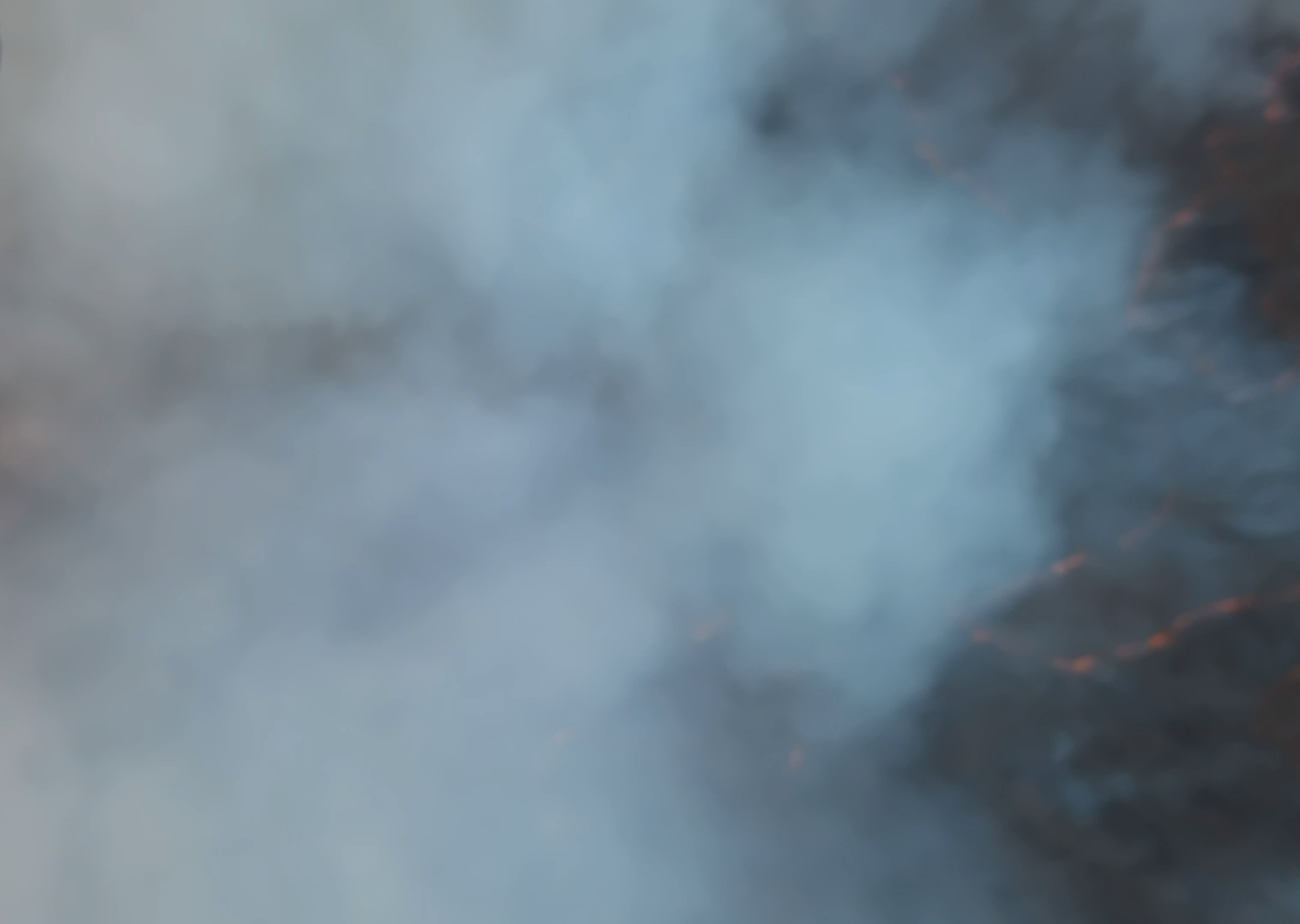}
        \caption{Barely visible fire in thick smoke.}
    \end{subfigure}
    \begin{subfigure}[b]{0.23\textwidth}
        \includegraphics[width=\textwidth]{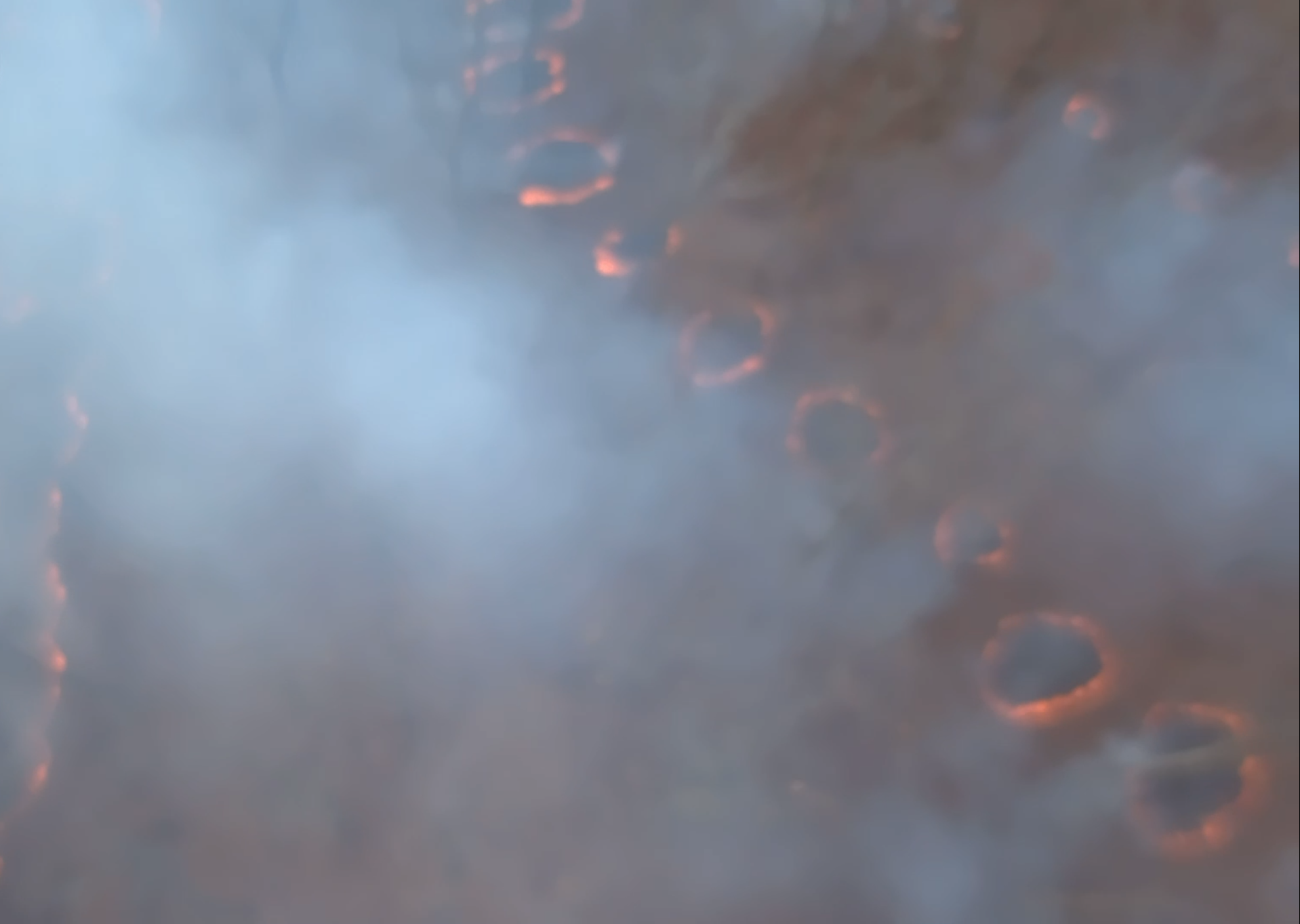}
        \caption{Visible fire in medium smoke thickness.}
    \end{subfigure}
    \begin{subfigure}[b]{0.217\textwidth}
        \includegraphics[width=\textwidth]{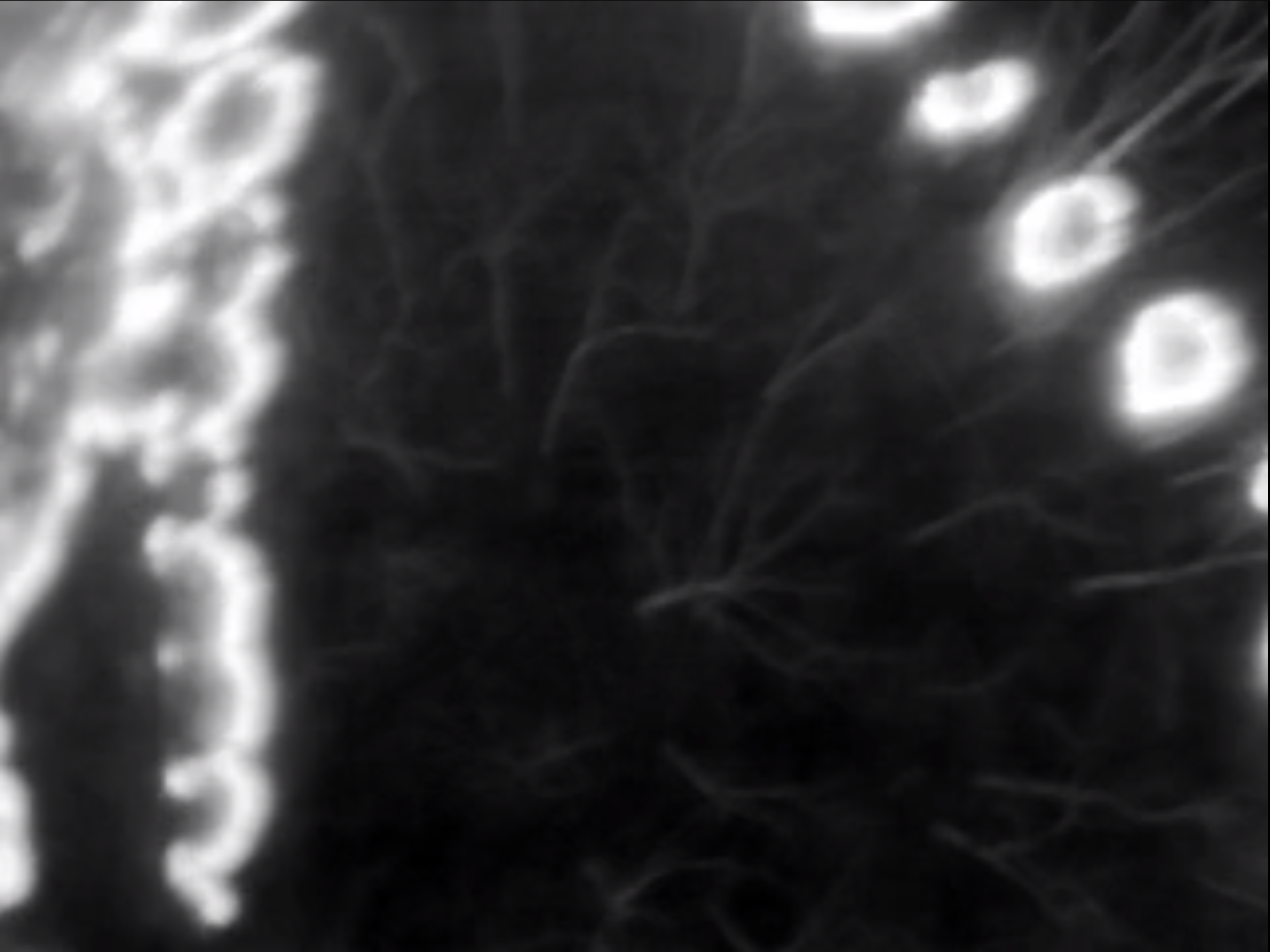}
        \caption{Seek thermal image comparison to (c).}
    \end{subfigure}
    \caption{\small{Samples RGB images from the WIT-UAV-ROS dataset collected from SGL 42 near Reade Township. The majority of RGB images in this dataset are occluded by heavy smoke, showing the need for thermal to penetrate. However, in areas of light smoke, RGB yields more diverse color features useful for asset identification.}}
    \label{fig:sample_rgb_images}
\end{figure}

\subsection{Data Description}
The three sites differ significantly in the amounts of vegetation and distribution of assets present.
We analyze this qualitatively and quantitatively below.

\textbf{Qualitative Data Description:}
The prescribed burn at SGL 174 near Rossiter, PA, recorded in the fall of 2021, features a dense forest.
About half the leaves had fallen off the trees, but dense branches resulted in traveling on foot with vehicles left behind.
As such, most labeled assets from this sequence are humans.
The dense branches create heavy occlusion resulting in relatively less visible assets aside from the clearance at the takeoff zone.
However, the visible moving humans are easy to distinguish.
A warm piece of cardboard was used as a takeoff pad, so a bright square lays on the ground in some of these sequences.

SGL 111 near Confluence, PA, recorded in late spring 2022, features an open clearing where most vehicles were parked.
The takeoff point was near the parking area, so many more vehicles are present for the Confluence sequences. 
Notably, in the Confluence sequences, it is possible to pick out the motion of human fire starters at the leading edge of the prescribed burn.
Moving vehicles are also easy to distinguish on roads.
The sharper image from the FLIR Boson yields more precision, but the absolute temperature of the Seek S304SP is a better indicator for fire thresholding.

Finally, SGL 42 which took place near Reade Township took place in fall 2022.
This burn produced significantly more smoke than the previous two burns.
Only the Seek S304SP was used on this burn.
Vehicles and people are highly visible thanks to the large open roads away from the trees.
The RGB camera data recorded at SGL 42 shows that color features are useful for identifying assets when not occluded by heavy smoke.
The heat signature from the Seek camera at this site shows much more prominently than in SGL 111, which, as we show later in Figure \ref{fig:ssd_false_positive_fire_comparision} increases false positives. 

\begin{figure}[!ht]
    \centering
    \includegraphics[scale=0.5]{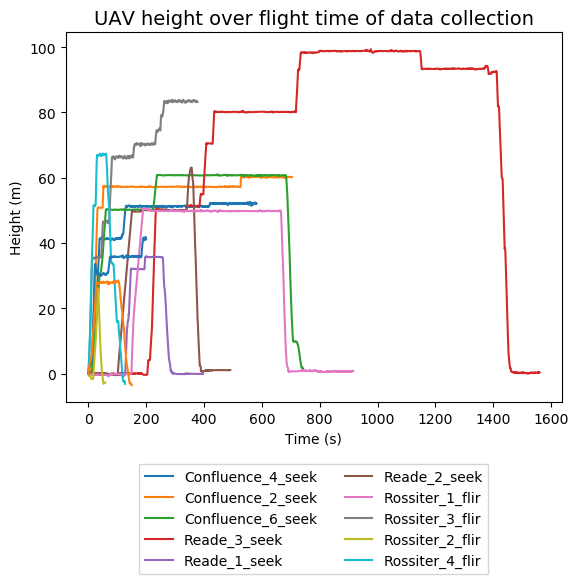}
    \caption{\small{UAV heights across flight time illustrate the diversity of our dataset.}}
    \label{fig:flyheights}
\end{figure}

\textbf{Quantitative Data Description:}
We analyze the Wildland-fire Infrared Thermal (WIT) dataset to illustrate the distribution of identified objects.
The dataset comprises a total of 6951 Wildland-fire images.
Out of these, 2062 images were manually labeled to identify humans and vehicles.
The distribution of labeled images with crew information is illustrated in Figure \ref{fig:data_distribution}.
The annotated dataset includes 5030 labeled vehicles and 1542 labeled human individuals.
Figure \ref{fig:data_distribution} depicts the distribution of labeled crews across the dataset.
The UAVs were flown at various heights ranging from 20 - 100 m.
The UAV flight heights were recorded throughout the runs and can be found in the accompanying WIT-UAS-ROS dataset.
Figure \ref{fig:flyheights} shows the flying heights for different data collection runs.

\section{Evalution and Analysis}

\subsection{Model Training}
\begin{figure}
    \centering
    \includegraphics[width=0.5\textwidth]{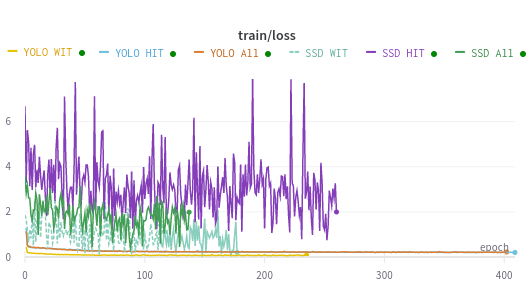}
    \caption{\small{Loss graphs during model training. Series are labeled with the model name and data it was trained on.}}
    \label{fig:loss_graph}
\end{figure}

\begin{figure}
    \centering
    \includegraphics[width=0.5\textwidth]{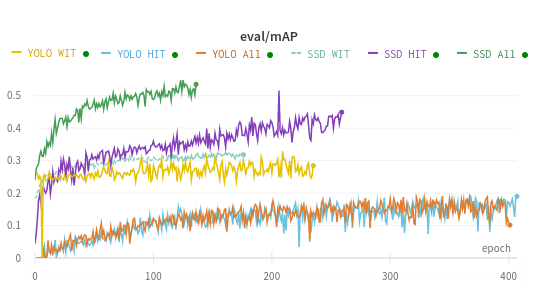}
    \caption{\small{Evaluation mAP on the validation set during model training. Series are labeled with the model name and data it was trained on.}}
    \label{fig:evaluation_mAP}
\end{figure}

\begin{figure}
    \centering
    \includegraphics[width=0.5\textwidth]{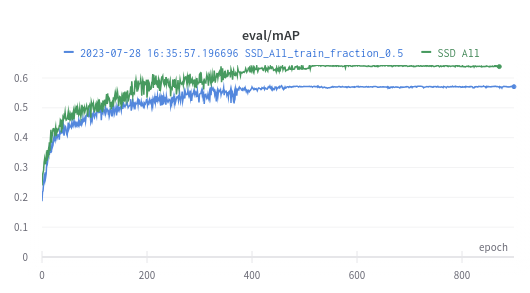}
    \caption{
    \small{
    Compares the best-performing model, SSD, when trained on 50\% of the HIT+WIT training data versus 100\% of the HIT+WIT training data.
    As expected, training with more data improves performance, here by 0.08 mAP points.
    }}
    \label{fig:eval_map_SSD_fractions}
\end{figure}

We trained YOLOv3 and SSD300 models on two different datasets: (1) HIT-UAV~\cite{hituav}, an urban thermal dataset, and (2) our WIT-UAS-Image fire dataset. We trained models until they plateaued for a max of 600 epochs, and the results are shown in Table~\ref{tab:training_result}.
We trained models trained only on HIT, only on WIT, and on both combined.
The model is optimized using Adam with learning rates of 0.001 and 0.0005 for YOLOv3 and SSD300. Models were trained with a batch size of 16.
Data augmentation included affine and minimal pixel-level transformations.
Training loss and validation mean average precision graphs are shown in Figures \ref{fig:loss_graph} and, \ref{fig:evaluation_mAP}.
To observe the effect of dataset size on mAP performance, we also compare the best-performing model, SSD300, when trained on only half vs the full amount of the HIT and WIT combined train dataset (Figure \ref{fig:eval_map_SSD_fractions}).

\begin{table*}[!ht]
    \centering
    \begin{subtable}[t]{\textwidth}
        \centering
        \begin{tabular}{|c|c|c|c|}
             \hline
             {\diagbox[]{Model}{Dataset}} & HIT & WIT & HIT\&WIT \\
             \hline \hline
             YOLOv3 & 0.103 / 0.028 / 0.067 / 0.197 & 0.025 / 0.010 / 0.046 / 0.081 & 0.115 / 0.029 / 0.067 / 0.210 \\
             SSD    & / / / 3.003 & / / / 1.212 & / / / 1.257 \\
             \hline
        \end{tabular}
        \caption{Training result format: class/object/IoU/overall loss}
        \label{tab:training_result}
    \end{subtable}
    \begin{subtable}[t]{\textwidth}
        \centering
        \begin{tabular}{|c|c|c|c|c|}
             \hline
             \multicolumn{2}{|c|}{{\diagbox[]{Model}{Dataset}}} & HIT & WIT & HIT\&WIT \\
             \hline \hline
             \multicolumn{1}{|c|}{\multirow{3}{*}{YOLOv3}} & HIT      & 0.20312 / 0.37046 / 0.12670 & 0.01045 / 0.10866 / 0.01581 & 0.12810 / 0.30333 / 0.06106 \\ \cline{2-2}
                                                           & WIT      & 0.00018 / 0.00473 / 0.00401 & 0.06103 / 0.16124 / 0.07693 & 0.01784 / 0.05313 / 0.03198 \\ \cline{2-2}
                                                           & HIT\&WIT & 0.20690 / 0.36084 / 0.12066 & 0.01747 / 0.11726 / 0.01793 & 0.12939 / 0.29731 / 0.06516 \\ \hline
             \multicolumn{1}{|c|}{\multirow{3}{*}{SSD}}    & HIT      & 0.444 / / & 0.054 / / & 0.330 / / \\ \cline{2-2}
                                                           & WIT      & 0.057 / / & 0.319 / / & 0.265 / / \\ \cline{2-2}
                                                           & HIT\&WIT & 0.484 / / & 0.317 / / & 0.511 / / \\ \hline
        \end{tabular}
        \caption{Validation result format: mAP/recall/f1}
        \label{tab:evaluation_result}
    \end{subtable}
    \begin{subtable}[t]{\textwidth}
        \centering
        \begin{tabular}{|c|c|c|c|c|}
             \hline
             \multicolumn{2}{|c|}{{\diagbox[]{Model}{Dataset}}} & HIT & WIT & HIT\&WIT \\
             \hline \hline
             \multicolumn{1}{|c|}{\multirow{3}{*}{YOLOv3}} & HIT      & 0.21908 / 0.39284 / 0.13465 & 0.01339 / 0.12247 / 0.01792 & 0.16661 / 0.34535 / 0.08326 \\ \cline{2-2}
                                                           & WIT      & 0.00059 / 0.00822 / 0.00637 & 0.06896 / 0.15951 / 0.07432 & 0.01497 / 0.04189 / 0.02586 \\ \cline{2-2}
                                                           & HIT\&WIT & 0.22554 / 0.38221 / 0.12809 & 0.01716 / 0.12524 / 0.01912 & 0.16896 / 0.33770 / 0.08281 \\ \hline
             \multicolumn{1}{|c|}{\multirow{3}{*}{SSD}}    & HIT      & 0.440 / / & 0.055 / / & 0.367 / / \\ \cline{2-2}
                                                           & WIT      & 0.138 / / & 0.310 / / & 0.239 / / \\ \cline{2-2}
                                                           & HIT\&WIT & 0.562 / / & 0.311 / / & 0.566 / / \\ \hline
        \end{tabular}
        \caption{Testing result format: mAP/recall/f1}
        \label{tab:testing_result}
    \end{subtable}
    \caption{\small{YOLOv3, SSD results on each dataset. For validation and testing, the datasets to the right of models indicate the data that they are trained on. Note that some of the results are left blank because unlike YOLO, SSD's loss and metric do not contain individual components.}}
\end{table*}

\subsection{Evaluation Process}

We then evaluated the trained model on HIT-UAV and WIT-UAS datasets.
The confidence threshold for candidate detection is 0.01, and the intersection over union(IoU) cutoff for valid detection is 0.5.
We apply a non-maximum suppression step with a threshold of 0.4 between the model output and benchmark.

Precision, recall, and f1 score are defined as: $\frac{True\;Positive}{True\;Positive + False\;Negative}$, $\frac {False\;Positive}{False\;Positive + True\;Negative}$, and $2 \cdot \frac{Precision \cdot Recall}{Precision + Recall}$.

\subsection{Detection Results}

\textbf{Quantitative Model Results:}
The results are shown in Table~\ref{tab:evaluation_result}, which produce the following insights:
\begin{itemize}
    \item YOLOv3 and SSD performed poorly when trained on only one data source but evaluated on the other, i.e. (1) trained on HIT and evaluated on WIT, (2) trained on WIT and evaluated on HIT.
          This is due to the reason that the WIT-UAS dataset contains thermal images of forests on fire, which is largely different from the urban environment of the HIT-UAV dataset, thus demonstrating a significant discrepancy between distributions of the data in each dataset, which results from the difference in environments where the data is collected.
    \item YOLOv3 and SSD demonstrate increases in mAP, recall, and f1 score when trained with both HIT and WIT datasets, albeit marginally.
          This shows that training with a combined dataset benefits detection, and a demand for more data in the domain of thermal images.
          Because of the nature of thermal images, objects at far in WIT appear inseparable with small blobs nearby of similar temperatures, which results in similar pixel values and more false positives.
\end{itemize}

\textbf{Qualitative Model Results:}

We visually compare samples from models trained only on urban data (HIT), models trained only on prescribed fire data (WIT), and models trained on both (HIT and WIT).

\begin{figure}[ht!]
    \centering
    \begin{subfigure}[b]{0.2\textwidth}
        \includegraphics[width=\textwidth]{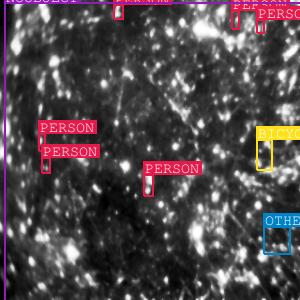}
        \caption{Sample 1: SSD trained on HIT detects false positives within the fire.}
    \end{subfigure}
    \begin{subfigure}[b]{0.2\textwidth}
        \includegraphics[width=\textwidth]{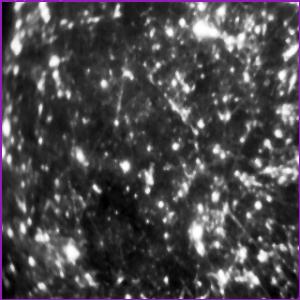}
        \caption{Sample 1: SSD trained on WIT shows no false positives in the fire.}
    \end{subfigure}
    \begin{subfigure}[b]{0.2\textwidth}
        \includegraphics[width=\textwidth]{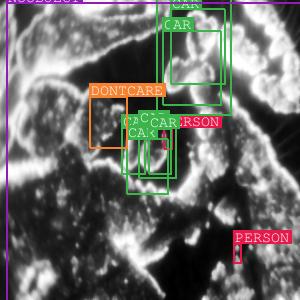}
        \caption{Sample 2: SSD trained on HIT shows false positives in the fire.}
    \end{subfigure}
    \begin{subfigure}[b]{0.2\textwidth}
        \includegraphics[width=\textwidth]{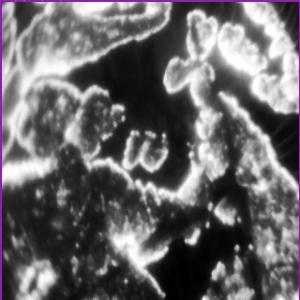}
        \caption{Sample 2: SSD trained on WIT shows no false positives in the fire.}
    \end{subfigure}
    
    \caption{\small{Example detections from SSD300 trained on HIT and WIT datasets inferred on Prescribed Fire Data. Notice that training on WIT significantly reduces the number of false positives.}}
    \label{fig:ssd_false_positive_fire_comparision}
\end{figure}

\begin{figure}[ht!]
    \centering
    \begin{subfigure}[b]{0.15\textwidth}
        \includegraphics[width=\textwidth]{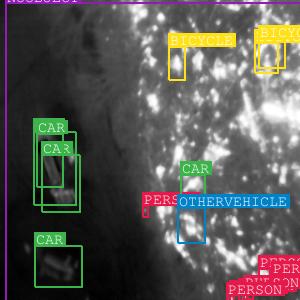}
        \caption{SSD trained on HIT.}
        \label{subfig:ssd_trained_hit_eval_wit}
    \end{subfigure}
    \begin{subfigure}[b]{0.15\textwidth}
        \includegraphics[width=\textwidth]{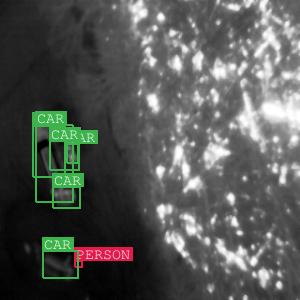}
        \caption{SSD trained on WIT.}
        \label{subfig:ssd_trained_wit_eval_wit}
    \end{subfigure}
    \begin{subfigure}[b]{0.15\textwidth}
        \includegraphics[width=\textwidth]{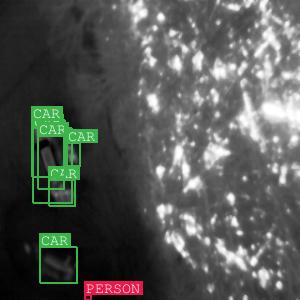}
        \caption{SSD trained on both.}
        \label{subfig:ssd_trained_on_both_eval_wit}
    \end{subfigure}
    \caption{\small{Detection samples on the WIT validation dataset with vehicles and fire. False positives were eliminated from including WIT data in training.}}
    \label{fig:hit_vs_wit_vs_all}
\end{figure}

We observe that models trained only on urban data  (HIT) but evaluated on wildland fire data (WIT) detected true positives for cars and people, but produced a significantly high number of false positives on the out-of-training-distribution fire data, detecting several fire hot spots as cars and people (Figure \ref{fig:ssd_false_positive_fire_comparision} and Figure \ref{subfig:ssd_trained_hit_eval_wit}).

We observe that models trained only on fire data (WIT) did not detect false positives in the fire (Figure \ref{fig:ssd_false_positive_fire_comparision} and Figure \ref{subfig:ssd_trained_wit_eval_wit}).

We observe that models trained on both urban and wildland fire data (HIT and WIT) were able to perform well on both sources of data, as expected because they are in-distribution; however, we observe that false positives on fire detections was significantly decreased (Figure \ref{subfig:ssd_trained_on_both_eval_wit}).

\subsection{Discussion}
We observe that even state-of-the-art single-frame object detection models trained on our data struggle with achieving high performance, which shows potential room for improvements in these algorithms.

Moreover, since the thermal images are extracted from sequences of images that are temporally close to each other, the objects will be easier to detect in a tracking manner, e.g. most cars and humans are dynamic across different frames above a relatively fixed background.
This suggests that further advancement of object tracking in thermal images could resort to video-based tracking algorithms.

While WIT alone reduced the false positive rates on fire, the models trained on HIT+WIT produced better performance overall. 
We observe that combining both urban (HIT) and wildland-fire (WIT) sources of thermal data leads to increases in performance.

    

\section{CONCLUSIONS}
Detection of assets of interest in wildland fire scenarios is integral to achieving wildfire safety monitoring. 
With the introduction of our Wildland-fire Infrared Thermal UAS dataset, we introduce for the first time a fire-environment detection dataset for people and vehicles.
We described the composition and uniqueness of our dataset, then trained and evaluated standard neural object detection models upon our data.
We showed that training on our dataset quantitatively improved performance on various metrics, and qualitatively reduced the number of false positives.
For future work, 
we recommend more data and more sophisticated models.
Gathering more data over prescribed burns in a similar manner to what was presented here would improve performance, but we also plan to develop a visually-realistic fire simulator to collect data. 
Additionally, more sophisticated detection models are recommended, for example, models that use video data to analyze motion or models that work well on small-object detection.

\addtolength{\textheight}{-12cm}   



\section*{ACKNOWLEDGMENT}



This work has been funded by the Department of Agriculture under award 20236702139073. Special thanks and appreciation goes to Arjun Chauhan, Kevin Gmelin, Sabrina Shen, Manuj Trehan, and Akshay Venkatesh for contribution designing the data collection platform and performing data collection.

\newpage

\bibliography{references}
\bibliographystyle{abbrv}

\end{document}